\documentclass[10pt,twocolumn,letterpaper]{article}

\usepackage{cvpr}
\usepackage{times}
\usepackage{epsfig}
\usepackage{graphicx}
\usepackage{amsmath}
\usepackage{amssymb}
\usepackage{times}  %Required
\usepackage{helvet}  %Required
\usepackage{courier}  %Required
\usepackage{url}  %Required
\usepackage{graphicx}  %Required
\usepackage{multirow}
\usepackage{tabularx} % for stretchable tables
\usepackage{caption}
\usepackage{xspace}
\usepackage{booktabs}
\usepackage{subfigure}
\usepackage{amsmath,amssymb} % define this before the line numbering.
\usepackage{enumitem}

\def \ie {{i.e.}\xspace}
\def \eg {{e.g.}\xspace}

\def \etal {{et al.}\xspace}

\newcommand{\para}[1]{\noindent \textbf{#1}}

\usepackage[pagebackref=true,breaklinks=true,letterpaper=true,colorlinks,bookmarks=false]{hyperref}

 \cvprfinalcopy % *** Uncomment this line for the final submission

\def\cvprPaperID{3517} % *** Enter the CVPR Paper ID here

\ifcvprfinal\fi
\begin{document}

%%%%%%%%% TITLE
\title{Point in, Box out: Beyond Counting Persons in Crowds}

\author{Yuting Liu$^1$,~~Miaojing Shi$^2$,~~Qijun Zhao$^1$,~~Xiaofang Wang$^2$ \\
	$^1$College of Computer Science, Sichuan University\\
	$^2$Univ Rennes, Inria, CNRS, IRISA \\
	 {\tt\small  yuting.liu@stu.scu.edu.cn; miaojing.shi@inria.fr; qjzhao@scu.edu.cn}
	%\and
}

\maketitle
\begin{abstract}
Modern crowd counting methods usually employ deep neural networks (DNN) to estimate crowd counts via density regression. Despite their significant improvements, the regression-based methods are incapable of providing
the detection of individuals in crowds. The detection-based methods, on the other hand, have not been largely explored in recent trends of crowd counting due to the needs for expensive bounding box annotations. In this work,
%%%%%%which is however important for person recognition and tracking.
%In this work, we revisit the detection-based methods in crowd counting,
%which have not been largely explored in recent trends due to the needs for expensive bounding box annotations.
we instead propose a new deep detection network with only point supervision required. It can simultaneously detect the size and location of human heads and count them in crowds. {We first mine
useful person size information from point-level annotations and initialize the pseudo ground truth bounding boxes. An online updating scheme is introduced to refine the pseudo ground truth during training; while a locally-constrained regression loss is designed to provide additional constraints on the size of the predicted boxes in a local neighborhood. In the end, we propose a curriculum learning strategy to train the network from images of relatively accurate and easy pseudo ground truth first.}
%1) pseudo ground truth bounding boxes are firstly initialized from point-level annotations and then iteratively updated during training;
%2) a locally-constrained regression loss is designed in point-supervised setting to let the predicted boxes have similar size to its local neighbors;
%3) %to cope with the varying density crowds,
%a curriculum learning strategy is proposed in the end to train the network from images of relatively accurate and easy pseudo ground truth first.
Extensive experiments are conducted in both detection and counting tasks on several standard benchmarks, \eg ShanghaiTech, UCF\_CC\_50, WiderFace, and TRANCOS datasets, and the results show the superiority of our method over the state-of-the-art.
%Particularly, in the counting task, we demonstrate that %our detection-based method performs close to density-based methods;
%by integrating our detection based method with recent density-based methods, we achieve significant improvement over the best so far.
\end{abstract}

	\section{Introduction}
\begin{figure}[t]
	\centering
	\includegraphics[width=1\columnwidth]{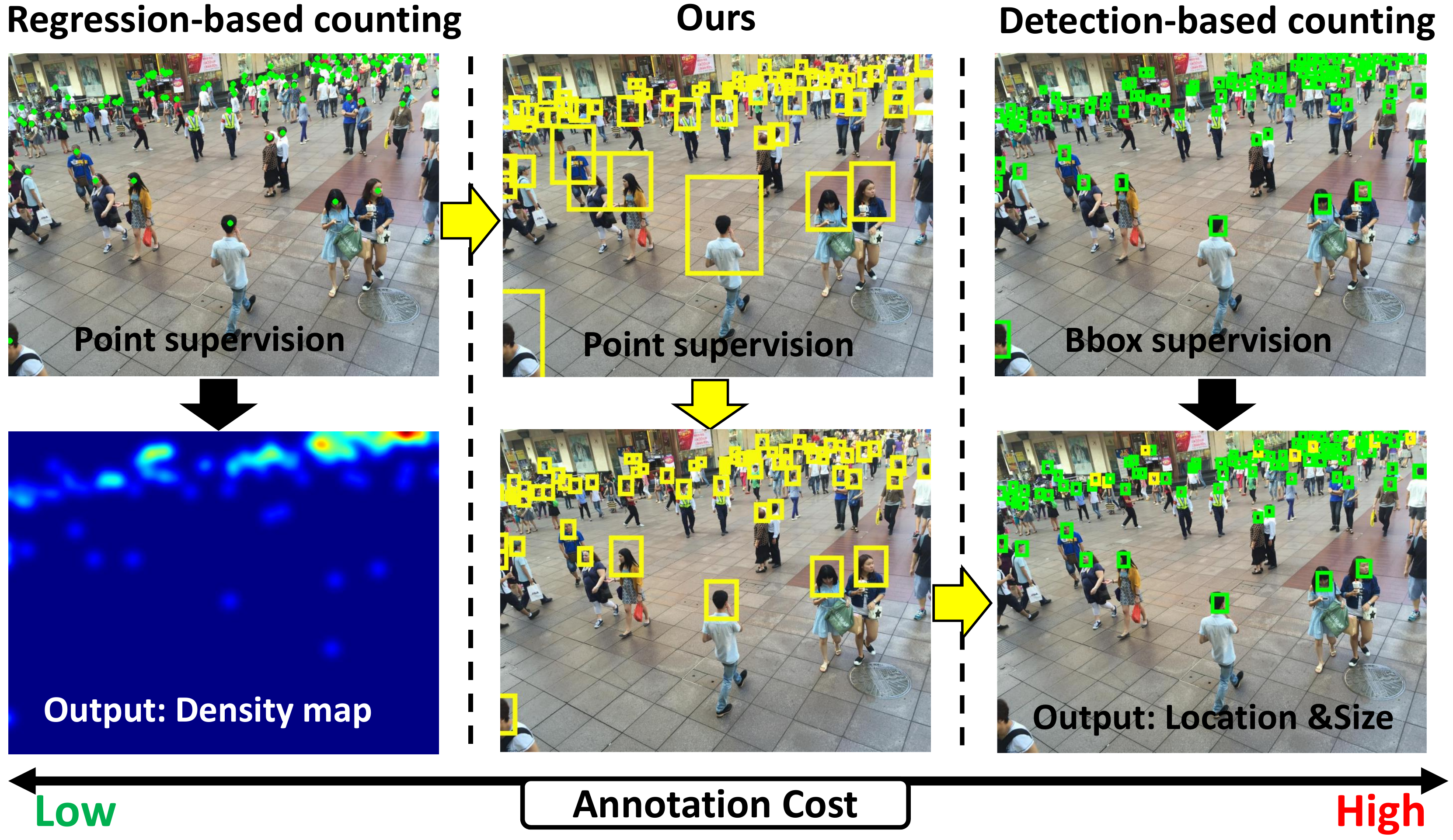}
	\caption{\small Comparison of our PSDDN with representative regression- and detection-based crowd counting methods regarding their annotation costs for input and their output information. 	}
	\label{Fig:motivation}
	\vspace{-0.1cm}
\end{figure}

Counting people in crowded scenes is a crucial component for a wide range of applications including video surveillance, safety monitoring, and behavior modeling. It is a highly challenging task in dense crowds due to heavy occlusions, perspective distortions, scale variations and varying density of people. Modern regression-based methods~\cite{onoro2016eccv,zhang2016cvpr,sindagi2017iccv, zhang2018wacv,liu2018cvpr,li2018cvpr,shi2019cvpr} cast the problem as regressing a density distribution map whose integral over the map gives the people count within that image (see Fig~\ref{Fig:motivation}: Left).
%This kind of point-level supervision is commonly provided by crowd counting datasets.
Owing to the advent of deep neural networks (DNN)~\cite{krizhevsky2012nips}, remarkable progress has been achieved in these methods.
%The density map for a training image is obtained via Gaussian blurring at every person head with only point supervision required, as is commonly provided in crowd counting datasets~\cite{zhang2016cvpr,zhang2015cvpr,chan2008cvpr}.
They do not require
annotating the bounding boxes but only the points of person heads at training. Yet, as a consequence, they
can not provide the detection of persons at testing, neither.

The detection-based methods, which cast the problem as detecting each individual in the crowds (see Fig~\ref{Fig:motivation}: Right), on the other hand, have not been largely explored in recent trends due to the lack of bounding box annotations. Liu~\etal~\cite{liu2018cvpr} have tried to manually annotate on partial of the bounding boxes in ShanghaiTech PartB (SHB) dataset~\cite{zhang2016cvpr} and train a fully-supervised Faster R-CNN~\cite{ren2015nips}. They combine the detection result with  regression result for crowd counting. Notwithstanding their efforts and obtained improvements, they did not report
results on datasets like SHA~\cite{zhang2016cvpr} and UCF\_CC\_50~\cite{idrees2013cvpr}, which have crowds on average five and ten times denser than that of SHB. 
%Although the detection-based methods cost huge annotation tasks, they can provide location and size information of persons which is very useful in person recognition~\cite{schroff2015cvpr}, tracking~\cite{rodriguez2011iccv}, and re-identification~\cite{liao2015cvpr}.

Annotating the bounding boxes of persons for training images can be a great challenge in crowd counting datasets. Meanwhile, knowing the person size and locations in a crowd at test stage is also very important; for example, in video surveilance, it enables person recognition~\cite{schroff2015cvpr}, tracking~\cite{rodriguez2011iccv}, and re-identification~\cite{liao2015cvpr}. Recently, some researchers~\cite{laradji2018eccv,idrees2018eccv} start to work on this issue with point supervision by employing segmentation frameworks~\cite{laradji2018eccv} or regressing the localization maps~\cite{idrees2018eccv} to simultaneously localize the persons and predict the crowd counts. Because they only use point-level annotations, they simply focus on localizing persons in the crowds, but do not consider predicting the proper size.

To be able to predict the proper size and locations of persons and meanwhile bypass the need for expensive bounding box annotations,
we introduce a new deep detection network using only point-level annotations on person heads (see Fig.~\ref{Fig:motivation}: Middle).
Although the real head size is not annotated, the intuition of our work is based on the observations that i) when two persons are close enough, their head distance indeed reflects their head size (similar to~\cite{zhang2016cvpr}); ii) due to the perspective distortion, person heads in the same horizontal line usually have similar size and gradually become smaller in the remote (top) area of the image. Both observations are common in crowd counting scenarios. They inspire us to mine useful person size information from head distances, and generalize a reliable point-supervised person detector with the help of head point annotations and size correlations in local areas.

To summarize, our work tries to tackle a very challenging yet meaningful task which is never handled by before; we propose a point-supervised deep detection network (PSDDN) for crowd counting which takes in cheap point-level annotations on person heads at training stage and produces out elaborate bounding box information on person heads at test stage. The contribution is three-fold:
\begin{itemize}[label=\textbullet, nolistsep, noitemsep]
\item We propose a novel online pseudo ground truth updating scheme which initializes the pseudo ground truth bounding boxes from point-level annotations (Fig.~\ref{Fig:motivation}: Middle top) and iteratively updates them during training (Fig.~\ref{Fig:motivation}: Middle bottom). The initialization is based on the nearest neighbor head distances.

 \item We introduce a novel locally-constrained regression loss in the point-supervised setting which encourages the predicted boxes in a local band area to have the similar size. The loss function is inspired from the perspective distortion impact on the person size in an image~\cite{shi2019cvpr,chan2008cvpr,zhang2015cvpr};

 \item We propose a curriculum learning strategy~\cite{bengio2009icml} to feed the network with training images of relatively accurate and easy pseudo ground truth first. The image difficulty is defined over the distribution of the nearest neighbor head distances within each image.
\end{itemize}
In extensive experiments, we show that (1) PSDDN performs close to those regression-based methods in crowd counting task on ShanghaiTech and UCF\_CC\_50 datasets; outperforms the state-of-the-arts by integrating with them. (2) In the mean time it produces very competitive results in person detection task on ShanghaiTech, UCF\_CC\_50, and WiderFace~\cite{yang2016cvpr} datasets. (3) We also evaluate PSDDN on the vehicle counting dataset TRANCOS~\cite{guerrero2015ibpra} to show its generalizability in other detection and counting tasks.

	\section{Related works}\label{Sec:RelatedWorks}
We present a survey of related works in three aspects: (1) detection-based crowd counting; (2) regression-based crowd counting; and (3) point-supervision.

\subsection{Detection-based crowd counting}
Traditional detection-based methods often employ motion and appearance cues in video surveillance to detect each individual in a crowd~\cite{viola2003ijcv,brostow2006cvpr,rabaud2006cvpr}. They are suffered from heavy occlusions among people. Recent methods in the deep fashion learn person detectors relying on exhaustive bounding box annotations in the training images~\cite{stewart2016cvpr,liu2018cvpr}. For instance, \cite{liu2018cvpr} have manually annotated the bounding boxes on partial of SHB and trained a Faster R-CNN~\cite{ren2015nips} for crowd counting. The annotation cost can be very expensive and sometimes impractical in very dense crowds. Our work instead uses only the point-level annotations to learn the detection model.
%It is carefully designed to detect objects in a crowd.

%: they did not annotate the bounding boxes in ShanghaiTech PartA and UCF\_FF\_50~\cite{idrees2013cvpr}, which are on average five and ten times denser than SHB, respectively. Our work instead learns a DNN model with only point supervision. It is carefully designed to detect objects in a crowd.

There are some other works particularly focusing on small object detection, \eg faces~\cite{hu2017cvpr,najibi2017iccv,bai2018cvpr}. \cite{hu2017cvpr} proposed a face detection method based on the proposal network~\cite{ren2015nips} while~\cite{najibi2017iccv} proposed to detect and localize faces in a single stage detector like SSD~\cite{liu2016eccv}. The face crowds tackled in these works are however way less denser than those in crowd counting works; moreover, these works are typically trained with bounding box annotations.

\subsection{Regression-based crowd counting}
Earlier regression-based methods regress a scalar value (people count) of a crowd~\cite{chan2008cvpr,chan2009iccv,idrees2013cvpr}. Recent methods instead regress a density map of a crowd; crowd count is obtained by integrating over the density map. Due to the use of strong DNN features,
%which have been demonstrated to be particularly good at solving problems ``locally" (density map) rather than ``globally" (scalar value),
remarkable progress has been achieved in recent methods~\cite{zhang2016cvpr,sam2017arxiv,sindagi2017iccv,zhang2018wacv,liu2018cvpr,liu2018bcvpr,ranjan2018eccv,idrees2018eccv,shi2019cvpr}.
%A density map provide much more information than a scalar value, and
More specifically,
%\cite{sam2017arxiv} extended the multi-column architecture in~\cite{zhang2016cvpr} and introduced a switch classifier to relay the crowd patches from images to the best DNN column.
\cite{sindagi2017iccv} designed a contextual pyramid DNN system. It consists of both a local and global context estimator to perform patch-based density estimation. \cite{liu2018bcvpr} leveraged additional unlabeled data from Google Images to learn a multi-task framework combining both counting information in the labeled data and ranking information in the unlabeled data. ~\cite{ranjan2018eccv} proposed an iterative crowd counting network
which first produces the low-resolution density map and then uses it to further generate the high-resolution density map.
%. It consists of multiple  two-branch unit where the first branch generates the low resolution density map while the second branch takes the output of the first branch and generates a high resolution density map.
Despite the significant improvements achieved in these regression-based methods, they are usually not capable of predicting the exact person location and size in the crowds.

%\subsection{Closest works to ours}
\cite{liu2018cvpr,laradji2018eccv,idrees2018eccv} are three most similar works to ours. \cite{liu2018cvpr} designed a so-called DecideNet to estimate the crowd density by generating detection- and
regression-based density maps separately; the final crowd
count is obtained with the guidance of an attention module. \cite{idrees2018eccv} introduced a new composition loss to regress both the density and localization maps together, such that each head center can be directly inferred from the localization map. \cite{laradji2018eccv} employed the hourglass segmentation network~\cite{ronneberger2015miccai} to segment the object blobs in each image for crowd counting; instead of using per-pixel segmentation labels, they use only point-level annotations~\cite{bearman2016eccv}.
Our work is similar to~\cite{liu2018cvpr} in the sense we both train a detection network for crowd counting; while \cite{liu2018cvpr} trained a fully-supervised detector using bounding box annotations, we train a weakly-supervised detector using only point-level annotations.  Our work is also similar to~\cite{idrees2018eccv,laradji2018eccv} where we all use point-level annotations; unlike our method, \cite{idrees2018eccv,laradji2018eccv} simply focus on person localization whereas we aim to predict both the localization and proper size of the person. Apart from all above, we also notice that we firstly evaluate the detection results on the dense crowd counting datasets \ie ShanghaiTech and UCF\_CC\_50.

%  for training
%Authors in~ and only reported results in datasets like SHB. \cite{idrees2018eccv} does not require bounding box annotations; on the other hand, it does not provide a way to obtain a proper bounding box around each head neither.

%We have mentioned that knowing the person locations in a crowd is very important to tasks like person recognition, tracking and re-identification. They require not just a dot on each person but rather a proper bounding box. Our method is built with only point-annotations at training and is able to predict proper bounding boxes at test.

\subsection{Point supervision}
{Point supervision} scheme has been widely used in human pose estimation to annotate
key-points of human body parts~\cite{johnson2010bmvc,ramanan2007nips,sapp2013cvpr}; while in object detection or segmentation, it has often been employed to reduce the annotation time~\cite{branson2011iccv,wah2011iccv,wang2014cviu,bearman2016eccv,papadopoulos2017iccv}.
For example,  Bearman~\etal~\cite{bearman2016eccv} conducted semantic segmentation by asking the annotators to click anywhere on a target object while Papadopoulos \etal~\cite{papadopoulos2017iccv} asked the annotators to click on the four physical points on the object for efficient object annotations.
The points can be collected either once offline~\cite{bearman2016eccv} or in an online interactive manner~\cite{branson2011iccv,wah2011iccv,wang2014cviu}.  We collect the points once and only use them at training time.
%Our work, for the first time, utilizes the point annotation of each head center to both detect and count persons in crowds.

 %\subsection{Curriculum learning}
%The curriculum learning paradigm was proposed
%by~\cite{bengio2009icml}, in which the model was learned gradually from easy samples to hard
%ones so as to increase the entropy of training. More specifically, in the computer vision domain, the concept of learning from easy to hard is focused on estimating the difficulty of images in order to decide the training order~\cite{lee2011cvpr,pentina2015cvpr,tudor2016cvpr,shi2016eccv,zhang2018cvpr}. \cite{tudor2016cvpr} estimated the image difficulty via analyzing low-level cues such as edges, segments, and
%objectness scores; \cite{shi2016eccv} proposed to feed images into a weakly-supervised learning loop in an order of object size, that is from images containing bigger objects down to smaller ones. \cite{zhang2018cvpr} proposed an easy-to-compute
%criterion named mean Accumulated Energy Scores (mEAS)
%to automatically measure the difficulty of an image. In this work, we decide the difficulty of images related to both the crowd density and pseudo ground truth accuracy. We learn our PSDDN with a specific order of training samples so that it can adjust to the varying density among samples.

	\section{Method}\label{Sec:Method}
\begin{figure*}[t]
	\centering
	\includegraphics[width=1\textwidth]{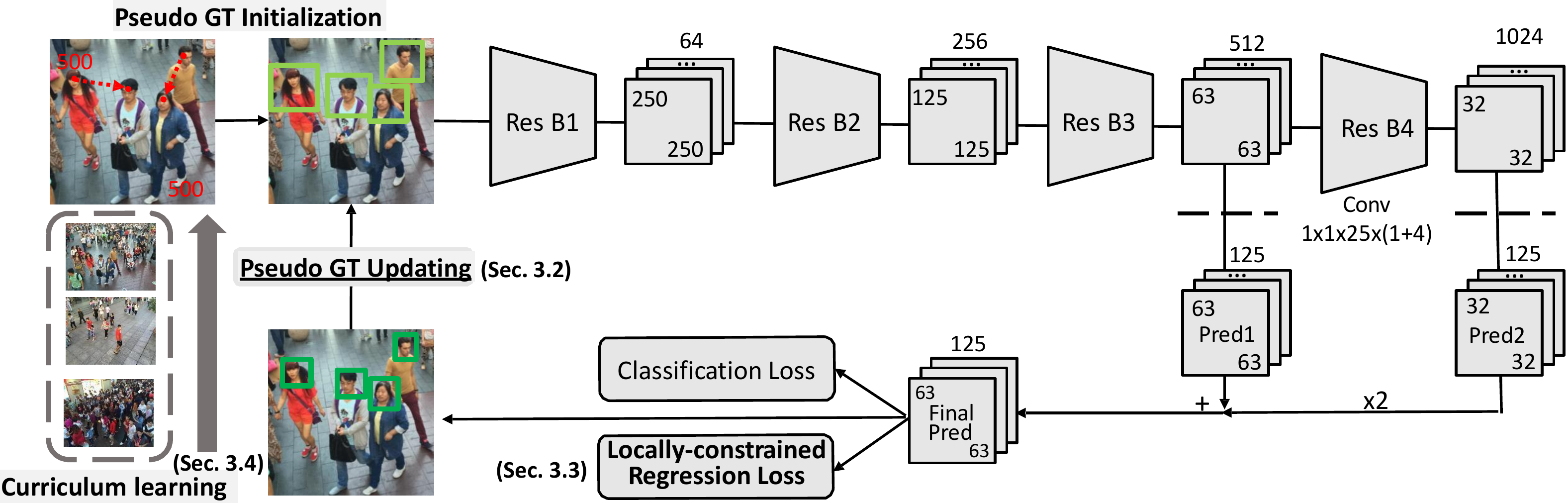}
	\caption{\small Network overview of PSDDN using only point-level annotations. Res B1 - B4 denotes the ResNet block adopted from ResNet-101~\cite{he2016cvpr}.
		Person detection is conducted on two scales after Res B3 and B4; their predictions (Pred1 and Pred2) are summed up to produce the final prediction (Final Pred). We propose an online pseudo ground truth updating scheme which includes pseudo ground truth initialization and updating;
		% from head points and iteratively updates them during training.
		a novel locally-constrained regression loss which encourages the predicted boxes in a local area to have the similar size. A curriculum learning strategy is proposed to train the network from images of relatively accurate pseudo ground truth first.
	}
	%B and D stands for the backbone and detection module, respectively; F is the feature map that connects B and D. Anchor boxes with different scale and aspect ratios are defined on F. Network is trained with both bounding box classification and regression loss, where the regression loss is to compare each box with ground truth boxes within a local band (red band on F, related to box row $i$). Ground truth are pseudo ground truth (green boxes), they are updated iteratively during training based on both box score and size (each box's width/height is compared to the distance from this head to its nearest neighbor, \eg the red dotted line on F). We propose a curriculum learning strategy to feed training samples to the network gradually according to their difficulty.
	\label{Fig:intro}
	\vspace{-0.1cm}
\end{figure*}
\subsection{Overview}
%Our model is based on the widely used anchor based detection framework, such as RPN~\cite{ren2015nips} and SSD~\cite{liu2016eccv}. The network architecture is shown in Fig.~\ref{Fig:intro} where we adopt our backbone from ResNet-101 with four 3-layer ResNet blocks (Res B1- B4)~\cite{he2016cvpr}.
%Likewise in~\cite{hu2017cvpr}, we select two convolutional layers after Res B3 and Res B4 as the detection layers, which are associated with different scales of anchors to predict detection.
%Each detection layer is followed by a convolutional (maybe not conv??) layer (Pred1/Pred2) that has the output of xx, where p is  xx and q is the anchor set size xxx. For each anchor, we predict
%4 offsets relative to its coordinates and 1 score for
%classification. Pred2 is up-sampled to the same resolution with Pred1 and added with it to produce the final predicted map (Final Pred). The multi-task loss of bounding box classification and regression is applied in the end.
%
%We extend the framework to point-supervised crowd counting with modules marked in bold in Fig~.\ref{Fig:intro}: a novel online ground truth (GT) updating scheme is firstly presented which incorporates initializing pseudo GT bounding boxes from point-level annotations and updating them during training.
%Afterwards, a \emph{locally-constrained regression loss} is specifically proposed for bounding box regression with point-supervision.
%In the end, we introduce a curriculum learning strategy to train our model from images of relatively accurate pseudo ground truth first.
Our model is based on the widely used anchor based detection framework, such as RPN~\cite{ren2015nips} and SSD~\cite{liu2016eccv}. The network architecture is shown in Fig.~\ref{Fig:intro} where we adopt our backbone from ResNet-101 with four ResNet blocks (Res B1- B4)~\cite{he2016cvpr}.
Likewise in~\cite{hu2017cvpr}, the outputs from Res B3 and Res B4 are taken to connect with two detection layers with different scales of anchors, respectively. The detection layer is a 1 x 1 convolutional layer that has the output of $N \times N \times T \times (1 + 4)$, where $N$ is the output length of feature maps and $T$ is the anchor set size (25 in our work). The aspect ratios of the predefined anchors are adapted from~\cite{ren2015nips} by referring to the centroid clustering of the nearest neighbor distance between person heads.
For each anchor, we predict
4 offsets relative to its coordinates and 1 score for
classification. Prediction \emph{Pred2} is up-sampled to the same resolution with \emph{Pred1} and added together to produce the final map \emph{Final Pred}. The multi-task loss of bounding box classification and regression is applied in the end.

We extend the framework to point-supervised crowd counting with modules marked in bold in Fig~.\ref{Fig:intro}: a novel online ground truth (GT) updating scheme is firstly presented which incorporates initializing pseudo GT bounding boxes from point-level annotations and updating them during training.
Afterwards, a \emph{locally-constrained regression loss} is specifically proposed for bounding box regression with point-supervision.
In the end, we introduce a curriculum learning strategy to train our model from images of relatively accurate pseudo ground truth first.
\subsection{Online ground truth updating scheme}\label{Sec:pseduo initliazation}
\emph{Pseudo ground truth initialization.} To train a detection network, we need to first initialize the ground truth bounding boxes from head point annotations. We follow the inspiration in~\cite{zhang2016cvpr} that the head size is indeed related to the distance between the centers of two neighboring heads in crowded scenes.
% They use it to estimate the Gaussian variance in density map generation; while
We use it to estimate the size of a bounding-box $g$ as the center distance $d(g,{\mathrm{NN}_g})$ from this head $g$ to its nearest neighbor ${\mathrm{NN}_g}$ (see Fig.~\ref{Fig:intro}: red dotted line). This makes a square bounding box; we find the corresponding anchor box that has the closest size to this square box as our initialization. We call the initialized bounding boxes pseudo ground truth. Some examples are shown in Fig.~\ref{Fig:motivation}: Middle top. The estimations in dense crowds (top) are close to the real ground truth while in sparse crowds (bottom) are often bigger.

\emph{Pseudo ground truth updating.} To train the detection network, we select positive and negative samples from the pre-defined anchors through their IoU (intersection-over-union) with the initialized pseudo ground truth. A binary classifier is trained over the selected positives and negatives so as to score each anchor proposal. Because the pseudo ground truth initialization is not accurate, we propose to iteratively update them to train a reliable object detector (see Fig.~\ref{Fig:motivation}). More formally, let $g^0$ denote an initialized ground truth bounding box at certain position of an image at epoch $0$.
%We select the positive and negative samples from the pre-defined anchors through their IoU (intersection-over-union) with $g^0$, \eg the positives and $g^0$ have IoU $> 0.8$ in our work.
%In the network detection module, %where $z$ signifies the down-sampled (by a factor of 8) position on the feature map of the detection module.
%some positive and negative samples are selected from pre-defined anchor boxes via their IoU (intersection-over-union) with $g^0$.
Over the positive samples of $g^0$, we select the highest scored one among those whose size (the smaller value of width or height) 
are smaller than $d(g, {\mathrm{NN_g}})$ to
%from head $g$ to its nearest head ${\mathrm{NN}_{g}}$, width the one with the highest classification score; if its
replace $g^0$ in the next epoch; \ie we denote it by $g^1$ at epoch 1. The anchor set is densely applied on each  detection layer, which guarantees that most pseudo ground truth can be updated with suitable predictions iteratively; if sometimes $g$ is too small to have positives, it will be simply ignored during training.

We notice that the classification loss we use is the same with~\cite{girshick2015iccv,ren2015nips}. Following~\cite{girshick2014cvpr,shrivastava2016cvpr}, we also apply the same online hard mining strategy regarding the positive and negative selections. Below we introduce our \emph{locally-constrained regression loss}.

\subsection{Locally-constrained regression loss}
We first refer to~\cite{girshick2014cvpr} for some notations in  bounding box regression. The anchor bounding box $a = (ax, ay, aw, ah)$ specifies the pixel coordinates of the center of $a$ together with its width and height in pixels. $a$'s corresponding ground truth $g$
%(refer to previous section for their correspondence)
is specified in the same way: $g = (gx, gy, gw, gh)$. %We drop the subscript position index for now.
The transformation required from $a$ to $g$ is parameterized as four variables $d_x(a)$,  $d_y(a)$,  $d_w(a)$,  $d_h(a)$. The first two specify a scale-invariant translation of the center of $a$, while the second two specify log-space translations of the width and height of $a$. These variables are produced by bounding box regressor; we can use them to transform $a$ into a predicted ground-truth bounding box $\hat g = (\widehat{gx}, \widehat{gy}, \widehat{gw}, \widehat{gh})$:
\begin{equation}\label{Eq:boundingbox}
 \begin{aligned}
\widehat{gx} = {aw} \cdot d_x(a) + ax,~
\widehat{gy} &= {ah}\cdot d_y(a) + ay\\
\widehat{gw} = {aw}\cdot \mathrm{exp}({d_w(a)}),~
\widehat{gh} &= {ah}\cdot \mathrm{exp}({d_h(a)})\\
 \end{aligned}
\end{equation}
The target is to minimize the difference between $g$ and $\hat g$.
\medskip

The ground truth $g$ in our framework is a pseudo ground truth: the center coordinates $gx$, $gy$ are accurate but $gw$, $gh$ are not. Based on this, we can not employ the original bounding box regression loss but instead we propose a \emph{locally-constrained regression loss}.

We first define a loss function $l_{xy}$ regarding the center distance between $g$ and $\hat g$:
\begin{equation}\label{Eq:Lregxy}
lxy = (gx - \widehat{gx})^2 + (gy - \widehat{gy})^2.
%lxy = (\frac{gx-ax}{aw} -  \frac{\widehat{gx}-ax}{aw})^2 + (\frac{gy-ay}{ah} - \frac{\widehat{gy}-ay}{ah})^2.
\end{equation}
With respect to the loss function on width and height, it is not realistic to directly compare between $g$ and $\hat g$. We rely on the observation (\emph{Observ}) that in a crowd image bounding boxes of persons along the same horizontal line should have similar size.
%while \textbf{ii)} boxes along the vertical line should have decreased size in general from the near front to the far end of the image.
This is due to the commonly occurred perspective distortions in crowd images: perspective values are equal in the same row, and decreased from the bottom to top of the image~\cite{chan2008cvpr,zhang2015cvpr,shi2019cvpr}.
As long as the camera is not severely rotated and the ground in the captured scene is mostly flat, the above observation should apply.
%Two assumptions are made to generate the perspective map: first, there is no in-plane rotation; second, the ground in the captured scene is flat. In standard crowd counting benchmarks, both assumptions holds for most of the images.
Hence, we propose to penalize the predicted bounding boxes $\hat g$ if its width and height clearly violate the \emph{Observ}.

Formally, denoting by $ g_{ij} = ({gx}_{ij}, {gy}_{ij},{gw}_{ij},{gh}_{ij})$ the pseudo ground truth at position $ij$ on the feature map,
%Notice that $i$ and $j$ are equivalent to ${gx}$ and ${gy}$; we keep both for our convenience.
we first compute the mean and standard deviation of the widths (heights) of all the bounding boxes within a narrow band area (row: $i-1: i+1$; column: $1:W$) on the feature map, $W$ is the feature map width. We use $G_i$ to denote the set of ground truth head positions within the narrow band related to $i$. The corresponding statistics are:
\vspace{-0.2cm}
\begin{equation}\label{Eq:Lregsigma}
\begin{aligned}
 \mu w_{i} &= \frac{1}{|G_i|}\sum_{{mn} \in G_i}{gw}_{mn}\\
 \sigma w_i&=\sqrt{\frac{1}{|G_i|}\sum_{{mn} \in G_i}({gw}_{mn} - \mu {w}_i)^2},\\
% \overline{gh}_{i} &= \frac{1}{3H}\sum_{k=i-1}^{i+1}\sum_{j=1}^H\widehat{gh}_{kj},\\
% \sigma(\widehat{gh}_i)&=\sqrt{\frac{1}{3H}\sum_{k=i-1}^{i+1}\sum_{j=1}^H(\widehat{gh}_{kj} - \overline{gh}_i)^2}\\
 \end{aligned}
% \vspace{-0.1cm}
\end{equation}
where $|G_i|$ signifies the cardinality of the set. $\mu {h}_{i}$ and $\sigma{h}_i$ can be obtained in the same way. We adopt a three-sigma rule: if the predicted bounding box width $\widehat{gw}_{ij}$ is larger than $\mu w_{i} + 3\sigma w_i$ or smaller than $\mu w_{i} - 3\sigma w_i$, it will be penalized; otherwise not. The loss function $lw_{ij}$ regarding the width of bounding box $\hat g_{ij}$ is thus defined as:
%\vspace{-0.1cm}
\begin{equation}\label{Eq:Lregwh}
lw_{ij} = \begin{cases} (\widehat{gw}_{ij} - (\mu w_{i} + 3\sigma w_i))^2 & \widehat{gw}_{ij} > \mu w_{i} + 3\sigma w_i \\ ((\mu w_{i} - 3\sigma w_i) - \widehat{gw}_{ij})^2 & \widehat{gw}_{ij} < \mu w_{i} - 3\sigma{w_i} \\0 & otherwise \end{cases}
%lw_{ij} = \begin{cases} (\ln{\frac{\widehat{gw}_{ij}}{aw}} - \ln{\frac{(\mu w_{i} + 3\sigma w_i)}{aw}})^2 & \widehat{gw}_{ij} > \mu w_{i} + 3\sigma w_i \\ (\ln{\frac{(\mu w_{i} - 3\sigma w_i)}{aw}} - \ln{\frac{\widehat{gw}_{ij}}{aw}})^2 & \widehat{gw}_{ij} < \mu w_{i} - 3\sigma{w_i} \\0 & otherwise \end{cases}
\end{equation}
$lh_{ij}$ can be obtained in a similar way. We do not require a restrict compliance with the \emph{Observ} in a local area, but instead design the narrow band and three-sigma rule for the tolerance of head size variation among individuals.

The overall bounding box regression loss $L\mathrm{reg}$ is:
\begin{equation}\label{Eq:Lreg}
L\mathrm{reg} = \sum_{ij \in G } \widetilde{lxy}_{ij} + \widetilde{lw}_{ij} + \widetilde{lh}_{ij},
\vspace{-0.1cm}
\end{equation}
where $G$ denotes the set of ground truth head points in one image. We add a tilde to each subloss symbol to signify that in real implementation the center coordinates, widths and heights of $g$ and $\hat g$ are normalized in a way related to the anchor box $a$ following the Eq. 6-9 in~\cite{girshick2015iccv}. 
%One can refer to many detection works~\cite{girshick2015iccv,ren2015nips} for these tricks.

%Similarly, we can obtain the corresponding statistics at line $i+1$: $\overline{gw}_{i+1}$, $\sigma(\widehat{gw}_{i+1})$, $\overline{gh}_{i+1}$, $\sigma(\widehat{gh}_{i+1})$; as well as line $i-1$: $\overline{gw}_{i-1}$, $\sigma(\widehat{gw}_{i-1})$, $\overline{gh}_{i-1}$, $\sigma(\widehat{gh}_{i-1})$.

\subsection{Curriculum learning}
Referring to Sec.~\ref{Sec:pseduo initliazation}: in very sparse crowds, the initialized pseudo ground truth are often inaccurate and much bigger than the real ground truth; on the other hand, in very dense crowds, the initializations are often too small and hard to be detected. Both cases are likely to corrupt the model and result in bad detection. Instead of training the model on the entire set once, we adopt a curriculum learning strategy~\cite{bengio2009icml,shi2016eccv,zhang2018cvpr} to train the model from images of relatively accurate and easy pseudo ground truth first.

%Conventional training scheme learns a model on the entire training set at the same time. Generalizing on either very sparse or very dense crowds can be difficult in the point-supervised setting. In sparse crowds, people tend to have big distance between each other which often results in bigger bounding box initialization compared to the real ground truth; while in the dense crowds, people tend to be very small and close to each other which can result in too small boxes to be detected. Both cases are likely to corrupt the model and result in bad detection. To avoid it, we introduce a curriculum learning strategy~\cite{bengio2009icml} to learn the model gradually from easy to hard.

% The former is due to the inaccurate bounding box initialization, which can be much bigger than the size of real ground truth in sparse crowds; while the latter is limited by the discriminating ability of the deep network. Both cases are likely to corrupt the model and result in bad detection. To avoid it, we introduce a curriculum learning strategy to learn the model gradually from easy to hard.
%The utilization of curriculum learning in vision tasks is usually focused on estimating the difficulty of images in order to decide the training order~\cite{lee2011cvpr,pentina2015cvpr,tudor2016cvpr,shi2016eccv,zhang2018cvpr}
Each pseudo ground truth $g$ is initialized with size $d(g,{\mathrm{NN}_g})$ (Sec.~\ref{Sec:pseduo initliazation}).
In a typical crowd counting dataset, very big or small boxes are only a small portion, most boxes are of medium/medium-small size, which are relatively more accurate and easier to learn.
% that is the distribution of each person's nearest neighbor distance $d(g,{\mathrm{NN}_g})$. We first compute the mean and
The mean $\mu$ and standard deviation $\sigma$ of $d(g,{\mathrm{NN}_g})$ can be computed over the entire training set. We therefore employ a Gaussian function $\Phi(d_g|\mu, \sigma)$ to produce scores for pseudo ground truth bounding boxes, such that the medium-sized boxes are in general assigned with big scores.
The mean score within an image is given by $\frac{1}{|G|} \sum_{g \in G} \Phi(d_g|\mu, \sigma)$, where $G$ denotes the bounding box set in the image. We define the training difficulty $\mathrm{TL}$ for an image as
\begin{equation}
\mathrm{TL} = 1- \frac{1}{|G|} \sum_{g \in G} \Phi(d_g|\mu, \sigma)
\end{equation}
 %$ 1- \frac{1}{|G|} \sum_{g \in G} \Phi(d_g|\mu, \sigma)$.
If an image contains mostly medium-sized bounding boxes, its difficulty will be small; otherwise, big.

Having the definition of image difficulty, we can split the training set $\mathcal I$ into $Z$ folds  $\mathcal I_1, \mathcal I_2,.., \mathcal I_Z$ accordingly. Likewise in~\cite{shi2016eccv, zhang2018cvpr}, we start by running PSDDN on the first fold $\mathcal I_1$ with images containing mostly medium-sized bounding boxes. Training on this fold will lead to a reasonable detection model.
After a couple of epochs running PSDDN on $\mathcal I_1$, the process moves on to the second fold $\mathcal I_2$, adding all its images into
the current working set $\mathcal I_1 \cup \mathcal  I_2$ and running PSDDN again.
%Instead of starting from scratch, we use $\mathcal B_1$ to update the pseudo ground truth in $\mathcal I_2$ such that the updated bounding boxes are tighter on persons.
The process will iteratively move on to the final fold $\mathcal I_Z$ and run PSDDN on the joint set $\mathcal I_1 \cup \mathcal  I_2 \cup ... \cup \mathcal  I_Z$. By the time it reaches $\mathcal  I_Z$ with images containing mostly super small/big bounding boxes, the model will already be very good and will do a much better job than training all the samples together from the very beginning. $Z$ is empirically chosen as 3 in our experiment.

	\section{Experiments}\label{Sec:Experiment}
%\holger{(Shortened this.)}
%The task of this paper is to do crowd counting, however we also provide the evaluation results for perspective estimations as a side task of PACNN in the Appendix
%\footnote{We will publish code and data to reproduce our experiments.}.
%\holger{Move publishing the code into the main text, not just as a footnote.}
We first introduce two crowd counting datasets and one face detection dataset. A vehicle counting dataset is also introduced to show the generalizability of our method. Afterwards, we evaluate our method on these datasets. %We offer more results in the supplementary material.
\subsection{Datasets}\label{Sec:Dataset}

\para{ShanghaiTech~\cite{zhang2016cvpr}.} It consists of 1,198 annotated images with a total of 330,165 people with head center annotations. This dataset is split into two parts: SHA and SHB. The crowd images are sparser in SHB compared to SHA: the average crowd counts are 123.6 and 501.4, respectively. Following~\cite{zhang2016cvpr}, we use 300 images for training and 182 images for testing in SHA; 400 images for training and 316 images for testing in SHB.
%\medskip
% with an average number of 50 pedestrians per image.
%\miaojing{we might mention perspective map here at some point}

\para{UCF\_CC\_50~\cite{idrees2013cvpr}.} It has 50 images with 63,974 head center annotations in total. The head counts range between 94 and 4,543 per image. The small dataset size and large variance make it a very challenging counting dataset.  We call it UCF for short. Following~\cite{idrees2013cvpr}, we perform 5-fold cross validations to report the average test performance.
%\medskip

\para{WiderFace~\cite{yang2016cvpr}.} It is one of the most challenging face datasets due to the wide variety of face scales and occlusion. It contains 32,203 images with 393,703 bounding-box annotated faces. The average annotated faces per image are 12.2.
40\% of the data are used as training, another 10\% form the validation set and the rest are the test set. The validation and test sets are divided into ``easy", ``medium", and ``hard" subsets. Test set evaluation has to be conducted by the paper authors. For convenience, we train all models on the train set and evaluate only on the validation set.

\para{TRANCOS~\cite{guerrero2015ibpra}.} It is a public traffic dataset containing 1244 images of different congested traffic scenes captured by surveillance cameras with 46,796 annotated vehicles.
The regions of interest (ROI) are provided for evaluation.
%\medskip

\subsection{Implementation details}\label{Sec:ExperimentalDetails}
To augment the training set, we randomly re-scale the input image by 0.5X, 1X, 1.5X, and 2X (four scales) and crop 500*500 image region out of the re-scaled output as training samples. Testing is also conducted with four scales of input and combined together. We set the learning rate as $10^{-4}$, with weight decay 0.0005 and momentum 0.9. Given the pseudo ground-truth and anchor bounding boxes during training, we decide positive samples to be those where IoU overlap exceeds 70\%, and
negative samples to be those where the overlap is below
30\%. We use a batch size of 12 images. In general, we train models for 50 epochs and select the best-performing epoch on the validation set. %The result is presented with a fixed confidence score for all datasets.
%We will publish our code to reproduce our experiments.

\subsection{Evaluation protocol}\label{Sec:Protocol}
We evaluate both the person detection and counting performance. For the counting performance, we adopt the commonly used mean absolute error (MAE) and mean square error (MSE)~\cite{sam2017arxiv,sindagi2017iccv,liu2018cvpr} to measure the difference between the counts of ground truth and estimation.

Regarding the detection performance, in the WiderFace
dataset, bounding box annotations are available for each face; a good detection $\hat g$ is therefore judged by the IoU overlap between the ground truth $g$ and detected bounding box $\hat g$, \ie $\mathrm{IoU}(g,\hat g) > 0.5$. In the ShanghaiTech and UCF\_CC\_50 datasets, we do not have the annotations of bounding boxes but only head centers. We define a good detection of $\hat g$ based on two criteria:
\begin{itemize}
	\item the center distance between the ground truth $g$ and detected $\hat g$ is smaller than a constant $c$.
	\item the width or height of $\hat g$ is smaller than $r*d(g, \mathrm{NN}_g)$, where $r$ is a constant.
\end{itemize}
%We will evaluate different $c$ and $r$ in the experiment.
%We choose by default $r = 1.5$ to allow some occlusion among person heads as well as some context into the detected boxes particularly in dense crowds; $c$ is set to $20$ (pixels).
$c$ is set to 20 (pixels) by default. As for $r$, there does not exist an exact selection of it since the real ground truth bounding boxes are not available. In dense crowds where persons are very close to each other or even occluded, $r$ could be a bit bigger than 1 to allow a complete detection around each head; while in sparse crowds, it is the opposite that $r$ should be smaller than 1. Building upon this, we choose $r$ by default as 0.8 for SHB and 1.2 for SHA and UCF. Different $c$ and $r$ will be evaluated in later sessions.

We compute the precision and recall by ranking our detected bounding boxes (good ones) according to their confidence scores. Average precision (AP) is computed eventually over the entire dataset.
 %\begin{equation}\label{Eq:MAE}
%\begin{split}
%&\mathrm{MAE} = \frac{1}{{N}}\sum\limits_{i = 1}^N {|{{Y^e_i} - {Y^g_i}}|},\\
%&\mathrm{MSE} = \sqrt{\frac{1}{{N}}\sum\limits_{i = 1}^N {({{Y^e_i} - {Y^g_i}})^2}},
%\end{split}
%\end{equation}
%where $Y^e_i$ and $Y^g_i$ denotes the estimated and the ground truth crowd counts, respectively. $Y^e_i$ is obtained by integrating the density map $D^e_i$ (\ref{Eq:DensityWeight}).
%Small MAE and RMSE values indicate good performance.
\begin{figure*}[t]
	\centering
	\includegraphics[width=1\textwidth]{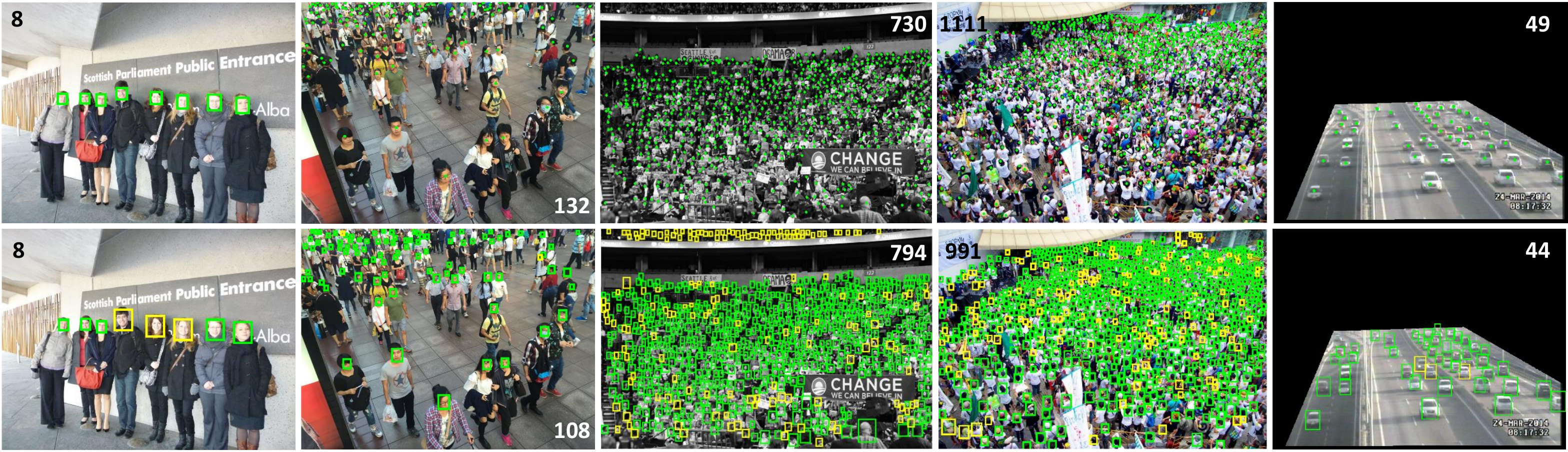}
	\caption{\small Examples from WiderFace, SHB, UCF, SHA, and TRANCOS datasets. The top row is test images with ground truth (bounding boxes or dots) while the bottom row is our detection. The numbers in images denote the ground truth and estimated counts, respectively. The green bounding boxes denote good detection while the yellows are not according to our evaluation protocol.}
	\label{Fig:result}
	\vspace{-0.0cm}
\end{figure*}

\subsection{Counting}\label{Sec:Ablation}
\para {ShanghaiTech~~}We first present an ablation study of PSDDN and then compare it with state-of-the-art.

\emph{Ablation study.} We present several variants (Pv0-Pv3) of PSDDN by gradually adding the proposed elements into the network. Referring to Sec.~\ref{Sec:Method}, we denote by Pv0 the model trained in a fully-supervised way using the fixed pseudo ground truth initialization and classic bounding-box regression as in~\cite{ren2015nips}; Pv1: the pseudo ground truth in Pv0 is iteratively updated; Pv2: the classic bounding box regression in Pv1 is upgraded to our new way; Pv3 (PSDDN): the curriculum learning strategy is adopted in Pv2.

The result is presented in Table~\ref{Tab:Counting} on both SHA and SHB. We take SHA as an example: the MAE for Pv0 starts from 168.6; it decreases to 104.7 for Pv1 and 89.8 for Pv2, respectively; finally, it reaches the lowest MAE 85.4 for Pv3, which is the full version of PSDDN. In the meantime, the MSE also significantly decreases from 268.3 of Pv0 to 159.2 of Pv3. We notice that the same observation goes with SHB as well. The result shows that each component of PSDDN provides a clear benefit in the overall system.

\begin{table}[t]
	\centering
	\small
	\begin{tabular}{|c|c|c|c|c|}
		\hline
		Dataset  & \multicolumn{2}{c|}{SHA}& \multicolumn{2}{c|}{SHB} \\
		\hline
		% \cline{2-7}
		Measures & MAE& MSE & MAE & MSE  \\
		\hline
		\hline
	Pv0   & 168.6 & 268.3 & 69.8 & 98.1  \\
	\hline
	Pv1    & 104.7 & 193.8 & 41.7 & 66.6 \\
	\hline
	Pv2    & 89.8 & 169.5 & 19.1 & 42.4 \\
	\hline
	Pv3(PSDDN)   & {85.4} & {159.2} & {16.1} & {27.9} \\
	\hline
%	PSDDN +  LSC18  & {77.6} & {125.6} & {10.8} & {16.9}\\
%	\hline
PSDDN +  \cite{li2018cvpr}  & \textbf{65.9} & \textbf{112.3} & \textbf{9.1} & \textbf{14.2} \\
	\hline
	\hline
	Li~\etal \cite{li2018cvpr} & \textbf{68.2} & 115.0 & \textbf{10.6} & \textbf{16.0}\\
\hline
Ranjan~\etal~\cite{ranjan2018eccv}& 68.5 & 116.2 & 10.7 & 16.0 \\
\hline
%	& 67.0  & 104.5 & 8.4& 13.6 & 258.4 & 344.9 \\
%	\hline
%	\cite{zhang2018wacv} & {86.8} & 139.2 & {16.2} & {25.8} & 314.9 & 424.8\\
%	\hline
Liu~\etal~\cite{liu2018bcvpr} & 73.6 & {112.0} & 13.7 & 21.4 \\
\hline
Liu~\etal~\cite{liu2018cvpr}  & - & - & {20.7} & 29.4\\
\hline
	DetNet in~\cite{liu2018cvpr}  & - & - & {44.9} & 73.2 \\
	\hline
%\hline
Sindagi~\etal~\cite{sindagi2017iccv} & {73.6} & \textbf{106.4} & {20.1} & {30.1} \\
\hline
Sam~\etal~\cite{sam2017arxiv}  & 90.4 & 135.0 & 21.6 & 33.4 \\
\hline
	\end{tabular}
	\caption{\small Crowd counting: ablation study of PSDDN (Pv0-Pv3 denote different variants of PSDDN) and its comparison with state-of-the-art on ShanghaiTech dataset. }
	%Times are reported in seconds per image and measured on ShanghaiTech PartB.  }
	\label{Tab:Counting}	
	%    \posttabspace
%	\vspace{-0.2cm}
\end{table}

\emph{Comparison with state-of-the-art.} We compare our work with prior arts~\cite{li2018cvpr,ranjan2018eccv,liu2018bcvpr,liu2018cvpr,sindagi2017iccv,sam2017arxiv}. It can be seen that our detection-based method PSDDN already performs close to recent density-based methods.
Furthermore, by combing our PSDDN result with~\cite{li2018cvpr} using the attention module in~\cite{liu2018cvpr}
%\footnote{We simply combine the regression result of \cite{li2018cvpr} with PSDDN using the trained module from~\cite{liu2018cvpr} at the inference time without finetuning.},
we show that the obtained result outperforms the state-of-the-art. For instance, on SHA, PSDDN + \cite{li2018cvpr} produces MAE  65.9 on SHA and 9.1 on SHB.
We notice two things: 1) we can obtain better counting results by adjusting the detection confidence scores;
on the contrary, we fix it with a high value (0.8) for all datasets to guarantee that the predictions are reliable at every local position;
2) the regression-based methods sometimes produce bad results in some local area of the image,
which can not be reflected in the MAE metric; there is another metric called GAME~\cite{guerrero2015ibpra} which is able to overcome this limitation. We will discuss later in TRANCOS dataset to show that our detection-based method is much better in the GAME metric.
%by 20+ points from \cite{li2018cvpr}. The lowest MAE and MSE reached by PSDDN+~\cite{li2018cvpr} are 44.4 and 78.1 on SHA, 8.3 and 12.6 on SHB, respectively.
We show some examples of PSDDN in~Fig.~\ref{Fig:result}.
%: the detection result is transformed to a density map and combined with the density output of \cite{li2018cvpr} using the attention module in \cite{liu2018cvpr}.

The notation ``DetNet" for \cite{liu2018cvpr} denotes the counting-by-detection result in it, where they annotate on partial of the bounding boxes in SHB and train a fully-supervised Faster R-CNN detector. PSDDN clearly outperforms the DetNet results.
%\miaojing{I commented out something about the curve below.}
%Since the code and data in \cite{liu2018cvpr} are not publicly available yet,
%Similar to the Fig.6 in \cite{liu2018cvpr}, we also draw a figure (Fig.~\ref{Fig:curve}(a)) to show the relationship between the
%predicted counts and ground-truth counts on the test
%set of SHB.
%It can be seen that the curve of PSDDN (green) fits the red line very well when the numbers of ground truth count are small. However,
%when the crowds get bigger, the crowd count by PSDDN becomes lower than that of the red line.
%Comparing our curve with that of \cite{liu2018cvpr} (DetNet), ours is closer to the ground truth distribution. It is in general difficult for detection-based methods to estimate crowd counts in dense crowds.
But we do not claim that point (weakly)-supervised learning is normally better than fully-supervised learning. Specifically for DetNet, they did not employ any of the data augmentation tricks as in PSDDN. The main limitation for fully-supervised detection methods in crowd counting lies in the large amount of bounding box annotations required. It can be unrealistic in very dense crowds. Our PSDDN instead provides an alternative way to conduct counting-by-detection with only point supervision; it performs very well in the evaluation of both counting and detection.
% \eg, \cite{liu2018cvpr} did not report experiment in SHA and UCF datasets, they are on average five and ten times denser than SHB. Our PSDDN instead provides an alternative way to conduct counting-by-detection with only point supervision; it performs very well in the evaluation of both counting and detection.
\medskip

\para{UCF\_CC\_50~~}It has the densest crowds so far in crowd counting task. We show in Table~\ref{Tab:Counting-UCF} that our PSDDN can still produce competitive result: the MAE is 359.4 while the MSE is 514.8.
In the detection session, we will show that despite the tiny heads in UCF, PSDDN is still able to produce reasonable bounding boxes on them (Fig.~\ref{Fig:result}: third column).

\begin{table}[t]
	\setlength{\tabcolsep}{2.3pt}
	\centering
	\small
	\begin{tabular}{|c|c|c|c|}
		\hline
		Counting & \multicolumn{3}{c|}{UCF} \\
		\hline
		% \cline{2-7}
		Measures & MAE& MSE & AP \\
		\hline
		\hline
		Li~\etal \cite{li2018cvpr}  & \textbf{266.1} &  \textbf{397.5}& - \\
		\hline
		Liu~\etal~\cite{liu2018bcvpr} & 279.6& 388.9 & - \\
		\hline
		Sindagi~\etal~\cite{sindagi2017iccv}  & 295.8 & 320.9 & -  \\
		\hline
		Sam~\etal~\cite{sam2017arxiv} & 318.1 & 439.2 & - \\
		\hline
		PSDDN   & {359.4} & {514.8} & 0.536 \\
		\hline
	\end{tabular}
	%	\begin{tabular}{|c|c|}
	%	\hline
	%	Detection & {UCF} \\
	%	\hline
	%	% \cline{2-7}
	%	Measures & AP \\
	%	\hline
	%	\hline
	%	PS-Faster R-CNN & 0.035\\
	%	\hline
	%		PSDDN   & \textbf{0.536} \\
	%	\hline
	%\end{tabular}
	\caption{Comparison of PSDDN with state-of-the-art on UCF dataset. MAE, MSE are reported for crowd counting while AP is reported for person detection.}  %PS-Faster R-CNN denotes Faster R-CNN~\cite{ren2015nips} trained with fixed pseudo ground truth for training. }
	\label{Tab:Counting-UCF}	
	%    \posttabspace
	\vspace{-0.2cm}
\end{table}

\subsection{Detection}
%\begin{table}[t]
%%	\setlength{\tabcolsep}{2.6pt}
%	\centering
%	\small
%	\begin{tabular}{|c||c|c|c|c|}
%		\hline
%		Dataset  & Pv0& Pv1 & Pv2 & Pv3 (PSDDN) \\
%		\hline
%		%SHA   & 0.147 & 0.350 & 0.400 & 0.402 \\ % eval_x1
%		%SHA   & 0.308 & 0.491 & 0.549 & 0.554 \\  % eval_x1.2
%		SHA   & 0.445 & 0.586 & 0.646 & 0.661 \\  % eval_x1.5
%		\hline
%		%SHB    & 0.132 & 0.426 & 0.653 & 0.711 \\
%		%SHB    & 0.269 & 0.526 & 0.686 & 0.734 \\
%		SHB    & 0.352 & 0.587 & 0.704 & 0.748 \\
%		\hline
%	\end{tabular}
%	\caption{\small Person detection: ablation study of PSDDN on ShanghaiTech (SHA and SHB) dataset. AP is reported. }
%	\label{Tab:Det-SH}	
%	%    \posttabspace
% %\vspace{-0.4cm}
%\end{table}

\begin{figure}[t]
	\centering
	\includegraphics[width=1\columnwidth]{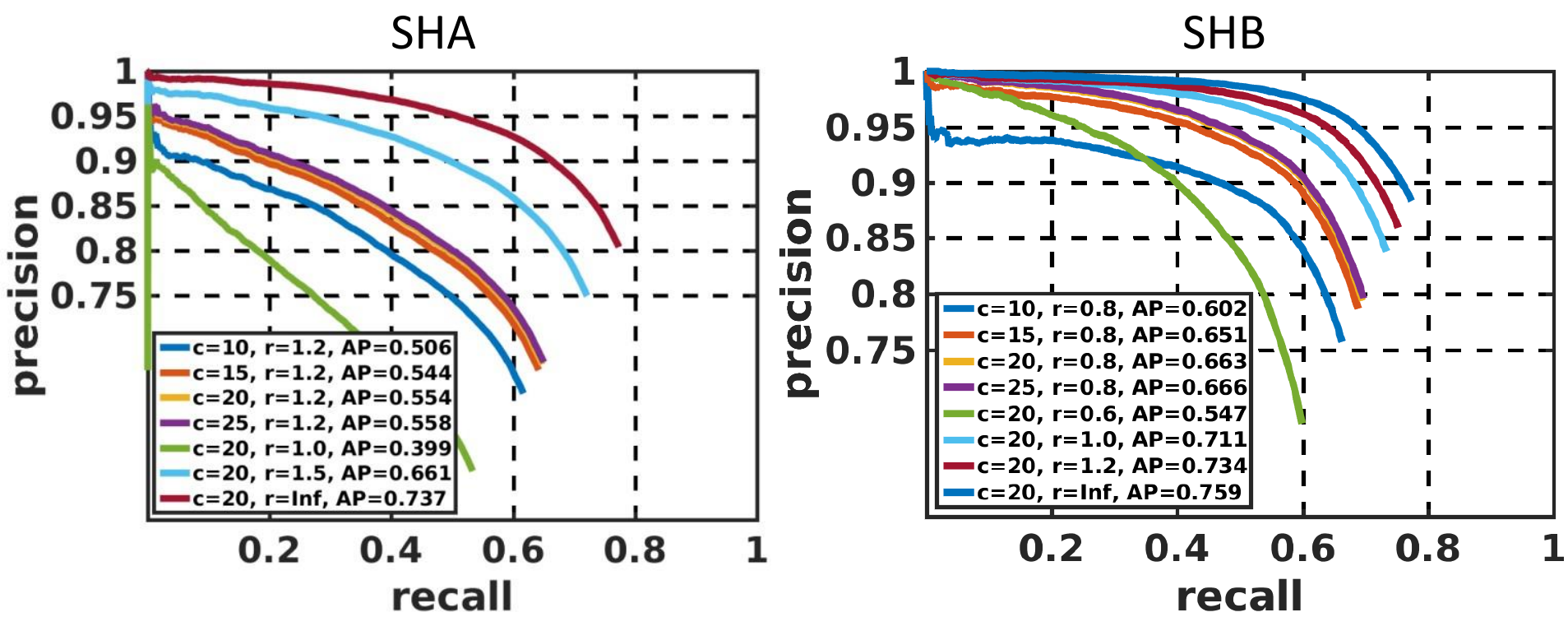}
%		\vspace{-0.6cm}
	\caption{\small Precision-recall curves with different $c$ and $r$.}
	\label{Fig:curve}
	\vspace{-0.3cm}
\end{figure}

\para{ShanghaiTech~~}In Fig.~\ref{Fig:curve}, we first present the precision-recall curves of different $c$ and $r$ (see Sec.~\ref{Sec:Protocol}) on SHA and SHB. The recall rates of different curves stop at some points as we fix the confidence score in the detection output. When we fix $r$, the AP improves with an increase of $c$; $c$ is chosen by default as 20 to apply a hard constraint on the center distance between the prediction and ground truth. On the other hand, when we fix $c$, the AP improves with an increase of $r$. As mentioned in Sec.~\ref{Sec:Protocol}, the crowds in SHA are much denser than in SHB, we choose by default $r = 1.2$ for SHA and $r = 0.8$ for SHB. We also present the result of $r = \infty$ which only cares the head center localizations (like in~\cite{laradji2018eccv,idrees2018eccv}): we get very good AP 0.737 and 0.759 for SHA and SHB, respectively.  \cite{laradji2018eccv,idrees2018eccv} did not present localization results in ShanghaiTech, we can not directly compare with them. But simply localizing the head centers is not enough for a detection task, we will further discuss in the WiderFace dataset where we have the real ground truth bounding boxes for evaluation.

%By increasing $c$ or $r$, we relax the constraint for a good detection, the AP result improves.  on the other hand, $r = 1.5$ as it allows some occlusion and a bit more context in the detected boxes rather than being too tight on those tiny heads (\eg in SHA, see Fig.~\ref{Fig:result}). This is not a critical issue for SHB as $r=1$ is already a loose constraint; we show that the AP results for $r=1.5$ and $r=1$ in SHB are very close (0.748 \emph{v.s.} 0.711).
%which indicates that the detected boxes in SHB are in general tightly bounded on people heads (see Fig.~\ref{Fig:result}).
%The reason we choose $r = 1.5$ by default is because it allow a bit more context  , which we believe is better for further recognition and tracking applications.
%We refer the readers to our supplementary material for the detailed results with $r = 1$.

Following the counting experiment, we also present the ablation study of PSDDN in detection. The result is shown in Table~\ref{Tab:Det-SH}: the AP on SHA is significantly increased from 0.308 for Pv0 to 0.554 for Pv3; the same  goes for SHB, where the AP is increased from 0.015 to 0.663 eventually. We notice we also tried to train a Faster R-CNN~\cite{ren2015nips} using the fixed pseudo ground truth, which is as low as in Pv0.

%\begin{table}[t]
%		\setlength{\tabcolsep}{3.5pt}
%	\centering
%	\small
%	\begin{tabular}{|c|c|c|c|c|c|c|}
%		\hline
%		 \multirow{2}{*}{Dataset} & \multirow{2}{*}{SHA}& \multirow{2}{*}{SHB} & \multirow{2}{*}{UCF} & \multicolumn{3}{c|}{WiderFace} \\
%		 \cline{5-7}
%	 & &  & & easy & medium & hard\\
%		\hline
%		\hline
%	%	PSDDN  & 0.402 & 0.645 & 0.421 & 0.539 &0.537 &0.316 \\
%	%	\hline
%			PSDDN  & \textbf{0.661} & \textbf{0.748} & \textbf{0.622} & 0.605 &0.605 &0.396 \\
%		\hline
%		Faster R-CNN  & 0.089 & 0.201 & 0.035 & 0.840 & 0.724 & 0.347 \\
%	\hline
%	HR17 & 0.396 & 0.239 & 0.261 & 0.925 & 0.910& {0.806} \\
%		\hline
%	NSC17  & - &-  &- & \textbf{0.931} & \textbf{0.921} & \textbf{0.845}\\	
%	\hline
%	\end{tabular}
%	\caption{\small Person detection: Comparison of PSDDN with state-of-the-art on datasets SHA, SHB, UCF, and WiderFace. Models of Faster R-CNN and HR17 on SHA, SHB and UCF employ the fixed pseudo ground truth for training, while on WiderFace employ the real ground truth.}
%	%Times are reported in seconds per image and measured on ShanghaiTech PartB.  }
%	\label{Tab:Detection}	
%	%    \posttabspace
%	% \vspace{-0.4cm}
%\end{table}

\begin{table}[t]
	\centering
	\small
	\begin{tabular}{|c||c|c|c|c|}
		\hline
		Dataset  & Pv0& Pv1 & Pv2 & Pv3 (PSDDN) \\
		\hline
		%SHA   & 0.147 & 0.350 & 0.400 & 0.402 \\ % eval_x1
		%SHA   & 0.308 & 0.491 & 0.549 & 0.554 \\  % eval_x1.2
		SHA   & 0.308 & 0.491 & 0.539 & 0.554 \\  % eval_x1.5
		\hline
		%SHB    & 0.132 & 0.426 & 0.653 & 0.711 \\
		%SHB    & 0.269 & 0.526 & 0.686 & 0.734 \\
		SHB    & 0.015 & 0.241 & 0.582 & 0.663 \\
		\hline
	\end{tabular}
	\caption{\small Person detection: ablation study of PSDDN on ShanghaiTech (SHA and SHB) dataset. AP is reported. }
	\label{Tab:Det-SH}	
	%    \posttabspace
	\vspace{-0.2cm}
\end{table}

%\begin{figure}[t]
%	\centering
%	\includegraphics[width=1\columnwidth]{fig/wider3}
%	\caption{\small Examples on UCF (left) and WiderFace (right) datasets. Note that in the left image, people on the upper floor are not annotated in ground truth but detected in PSDDN (\eg yellow ones).
%	}
%	\label{Fig:Wider}
%		\vspace{-0.1cm}
%\end{figure}

\begin{table}[t]
		\setlength{\tabcolsep}{2.2pt}
	\centering
	\small
	\begin{tabular}{|c||c|c|c|c|}
		\hline
		 \multirow{2}{*}{Methods} & \multirow{2}{*}{Annotations} & \multicolumn{3}{c|}{WiderFace} \\
		 \cline{3-5}\
	  & & easy & medium & hard\\
	  \hline
	  \hline
		Avg. BB & points(test)+ mean size & 0.002 & 0.083 & 0.059\\	
	%	\hline
	%		BB size: $d_{3\mathrm{NN}}$ & point(test) & {0.931} & {0.921} & {0.845}\\	
		\hline
		FR-CNN (ps) & points(train) + mean size  & 0.008 & 0.183 & 0.108 \\
		\hline
				FR-CNN (fs) & bounding boxes (train) & \textbf{0.840} & \textbf{0.724} & 0.347 \\
		\hline
			PSDDN  & points(train) &{0.605} &{0.605} &\textbf{0.396} \\
		\hline
	\end{tabular}
	\caption{\small Person detection on WiderFace. ``Annotations" denotes different levels of annotations employed in the methods. ``mean size" refers to the mean ground truth bounding box size over the training set while ``point(test)" specifically denotes that the bounding box centers are known for test. Avg. BB adds bounding boxes at each test point using the mean size. FR-CNN: Faster R-CNN.  }
	%Times are reported in seconds per image and measured on ShanghaiTech PartB.  }
	\label{Tab:Wider}	
	%    \posttabspace
	 \vspace{-0.1cm}
\end{table}
 \medskip

\para{UCF\_CC\_50~~}Table~\ref{Tab:Counting-UCF} shows the detection performance of PSDDN on UCF. In this dataset with very dense crowds, our method still achieves the AP of 0.536. An example is shown in Fig.~\ref{Fig:result}: third column. We refer the readers to those people sitting in the upper balcony (\eg yellow ones): they are not annotated as ground truth but detected by PSDDN.
\medskip

\para{WiderFace~~} \label{Sec:Widerface}
WiderFace is a face detection dataset, its crowd density is less denser than that in a typical crowd counting dataset; we report results in Table~\ref{Tab:Wider} to show the generalizability of our method. It can be seen using only point-level annotations PSDDN still manages to achieve AP 0.605, 0.605, 0.396 on the easy, medium, and hard set.

%The result on the hard set (overall set) is already close to that of Faster R-CNN.
\emph{Comparison to others.} Since we have the bounding box annotations available for both training and test in WiderFace, we try to compare PSDDN with~\cite{laradji2018eccv,idrees2018eccv,liu2018cvpr}. \cite{laradji2018eccv,idrees2018eccv} predicts either localization maps or segmentation blobs for both object localization and crowd counting. Predicting the exact size and shape of the object is not considered necessary for crowd counting in their works. However, we argue that it is important to object recognition and tracking. We assume there exists another method that can correctly localize every head center at test (better than any of \cite{laradji2018eccv,idrees2018eccv}), bounding boxes are added in a post-processing way using the mean ground truth size from the training set. It is denoted as \emph{Avg.BB} in Table~\ref{Tab:Wider}. The results are very low. We notice that we also tried to add the boxes in a similar way to our pseudo ground truth initialization at each test point, the APs are also very low. This demonstrates that it is not straightforward to add bounding boxes on top of the head point localization results. We also compare PSDDN with Faster R-CNN \cite{ren2015nips} using two different  levels of annotations in Table~\ref{Tab:Wider}: F\emph{R-CNN(ps)} and \emph{FR-CNN(fs)}. First, we use the head point annotations together with the mean ground truth size to generate  bounding boxes for training; it performs much worse than our PSDDN. Next, we follow~\cite{jiang2017fg} to use the manually annotated bounding boxes to train Faster R-CNN, which is analogue to the DetNet in~\cite{liu2018cvpr}.
PSDDN performs lower AP than FR-CNN(fs) on the easy and medium set but higher AP on the hard set. We point out that, many faces are well covered by the detection of PSDDN but not taken as good ones (yellow ones in Fig.~\ref{Fig:result}: first column) only because of their low IoU with the annotated ground truth. We believe this has displayed some potential for future improvement.
\medskip

\para{TRANCOS~~}We evaluate PSDDN on TRANCOS dataset to test its generalizability, though it is proposed for person detection and counting. The Grid Average Absolute Error (GAME) is used to evaluate the counting performance. We refer the readers to~\cite{li2018cvpr,guerrero2015ibpra} for the definition of GAME(L) with different levels of $L$. For a specific $L$, GAME(L) subdivides the image using a grid of $4^{L}$ non-overlapping regions, and the error is computed as the sum of the mean absolute errors in each of these regions. When L = 0, the GAME is equivalent to the MAE metric. We present the result of our PSDDN in Table~\ref{Tab:TRANCOS} where we obtain 4.79, 5.43, 6.68 and 8.40 for GAME0, GAME1, GAME2 and GAME3, respectively. Comparing our method with the state-of-the-art, PSDDN outperforms the best regression-based method~\cite{li2018cvpr} on GAME1, GAME2 and GAME3 and is competitive with it on GAME0. Unsurprisingly, the GAME theory is designed to penalize those predictions
with a good MAE but a wrong localization of the objects.
Our method produces good results on both overall vehicle counting and local vehicle localization/detection. The AP result of PSDDN for detection is 0.669 with $r=1$.
\begin{table}[t]
	\setlength{\tabcolsep}{2.1pt}
	\centering
	\small
	\begin{tabular}{|c|c|c|c|c|c|}
		\hline
		% \cline{2-7}
		Methods & GAME0& GAME1 &GAME2 & GAME3 & AP \\
		\hline
		\hline
		Victor~\etal~\cite{lempitsky2010nips} & 13.76 & 16.72 &  20.72 & 24.36 & - \\
		\hline
		Onoro~\etal~\cite{onoro2016eccv} &10.99 & 13.75 & 16.09 & 19.32 & -\\
		\hline
	%	\cite{zhang2017iccv} 4.21 & - & - & - & - &  \\
%		\hline
		Li~\etal~\cite{li2018cvpr} & \textbf{3.56} &  {5.49} & 8.57 & 15.04 & - \\
		\hline
		PSDDN   & {4.79} & \textbf{5.43} & \textbf{6.68} & \textbf{8.40} & \textbf{0.669}\\
		\hline
		%	PSDDN + \cite{zhang2018wacv}  & {77.6} & {125.6} & {10.8} & {16.9} & - & -\\
		%		\hline
	\end{tabular}
	\caption{\small Results on TRANCOS dataset.  }
	%Times are reported in seconds per image and measured on ShanghaiTech PartB.  }
	\label{Tab:TRANCOS}	
	%    \posttabspace
	\vspace{-0.2cm}
\end{table}

	\section{Conclusion}
In this paper we propose a point-supervised deep detection network for person detection and counting in crowds. Pseudo ground truth bounding boxes are firstly initialized from the head point annotations, and updated iteratively during the training. Bounding box regression is conducted in a way to compare each predicted box with the ground truth boxes within a local band area. A curriculum learning strategy is introduced in the end to cope with the density variation in the training set. Thorough experiments have been conducted on several standard benchmarks to show the efficiency and effectiveness of PSDDN on both person detection and crowd counting. Future work will be focused on further reducing the supervision in this task.
\medskip 

\noindent \textbf{Acknowledgments.} This work was supported by National Key Research and Development Program of China (2017YFB0802300), National Natural Science Foundation of China (61828602 and 61773270).

{\small
\bibliographystyle{ieee}
\bibliography{cvpr2019}
}

\end{document}